# Enabling Machine Learning across Heterogeneous Sensor Networks with Graph Autoencoders

Johan Medrano[1][0000-0002-7558-2071] and Fuchun Joseph Lin[2]

[1]Institut National des Sciences Appliquées, Toulouse, France
[2]Department of Computer Science, National Chiao Tung University, Hsinchu, Taiwan

**Abstract.** Machine Learning (ML) has been applied to enable many life-assisting applications, such as abnormality detection and emdergency request for the solitary elderly. However, in most cases machine learning algorithms depend on the layout of the target Internet of Things (IoT) sensor network. Hence, to deploy an application across Heterogeneous Sensor Networks (HSNs), i.e. sensor networks with different sensors type or layouts, it is required to repeat the process of data collection and ML algorithm training. In this paper, we introduce a novel framework leveraging deep learning for graphs to enable using the same activity recognition system across HSNs deployed in different smart homes. Using our framework, we were able to transfer activity classifiers trained with activity labels on a source HSN to a target HSN, reaching about 75% of the baseline accuracy on the target HSN without using target activity labels. Moreover, our model can quickly adapt to unseen sensor layouts, which makes it highly suitable for the gradual deployment of real-world ML-based applications. In addition, we show that our framework is resilient to suboptimal graph representations of HSNs.

**Keywords:** Graph Autoencoders, Heterogeneous Sensor Networks, Smart Homes

## 1  Introduction

The development of networking technologies and the advance in embedded computing enable the widespread deployment of IoT sensor networks. In this context, pervasive sensing and actuation with ubiquitous Internet of Things (IoT) tends to be a natural direction. The development of machine learning algorithms with pervasive sensing enables applications that can significantly enhance our daily lives. Some outstanding examples include abnormality detection in the routines of solitary elderly persons [1], assisted living for people with dementia [2, 3], early detection of Parkinson disease [4].

To perform these tasks, machine learning algorithms can take advantage of a large amount of data available for training. However, in most cases these algorithms are strongly tied to the physical layout of the sensors [5]. This makes their general applicability on real-world applications doubtful. First, due to the dependence on the structure of the sensor network, the deployment of applications to other sensor networks is impaired as it requires the collection of new data and the training of a new machine learn-ing model. Consequently, each deployment requires a repetition of the same effort. Second, when the model is trained online with the collected data, the application requires a significant amount of time before achieving decent performances. To reduce the deployment overhead and enable large-scale applications, there is an urgent need for a solution allowing the use of the same machine learning model across HSNs.

The existing frameworks rely on complex methods often tied with the adapted machine learning model. With the objective of proposing a simpler and more generic approach to adapt machine learning models, in particular classifiers, across HSNs, we introduce a novel framework in this paper. In the proposed architecture as depicted in Fig. 1, a first component, *Graph Autoencoder* (GAE), handles the task of domain adaptation across HSNs while a second component, the *structure-dependent classifier*, ensures the classification task. This design allows the application of fundamentally different classifiers on the top of the same cross-network adapter model.

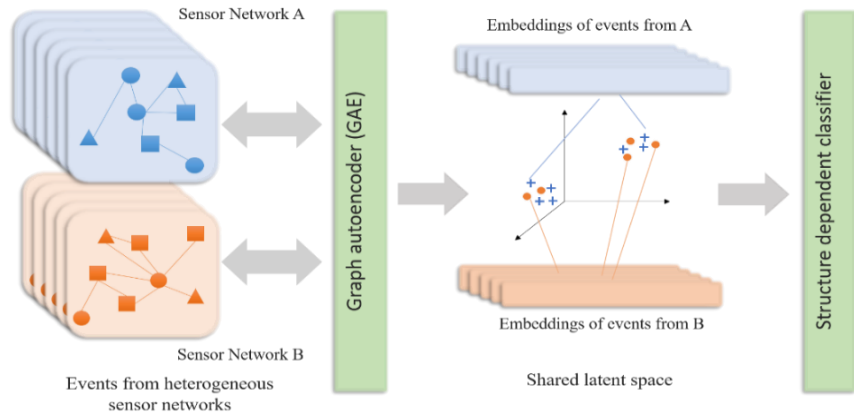

**Fig. 1.** Cross-network adaptation framework based on a graph autoencoder

We apply the proposed framework to the use case of *activity recognition in smart homes*. Smart homes typically have different layouts and sensors. In practical use cases such as abnormality detection in the routines of solitary elderly people, the well-being of the inhabitant would depend on the deployment of a well-trained model for accurate activity recognition. Focused on this critical use case, we train our encoder with data gathered in CASAS datasets [6] that were collected in smart homes with heterogeneous sensor layouts. The contributions of this paper are as follows:

- We propose a simple graph representation for HSN deployed in smart homes to enable the application of deep learning methods to the HSNs.
- We introduce a novel graph autoencoder architecture, which uses graph convolutional networks [7] together with differentiable pooling [8] to project data coming from HSNs into a latent space with fixed dimensions,
- We leverage associative domain adaptation [9] on our graph autoencoders to minimize the discrepancy of latent representations issued from different sensor networks and thus make the latent space sharable between sensor networks.

The rest of this paper is organized as follows. In Section 2, we introduce related methods which have been used for cross-domain adaptation or learning transfer across sensor networks. Section 3 then provides necessary prerequisites and describes the proposed framework. In Section 4, we introduce the experimental setup, subsequent parameters settings for our model and collected results. We discuss in Section 5 the results and parameters influence. Finally, we conclude and open future directions in Section 6.

## 2 Related Work

Several architectures have been proposed to perform heterogeneous domain adaptation (HDA), which aims at applying to a target domain the knowledge acquired from a source domain. Methods can be divided in two categories: *feature remapping* methods, focusing on finding a mapping between the features in source and target domains, and *latent space transformation* methods, constructing projections of data from source and target domains in a shared latent space.

The key principle of *feature remapping* is to find an optimal mapping between features of heterogeneous domains where the features in target domain can be associated to a single feature or to a combination of features in the source domain. Given the sensor readings of source and target domains and the labels for source domain, Hu and Yang [10] assumed that feature representations in source and target domains are similar. Based on this assumption, they constructed a *translator* that automatically finds a mapping between source and target features. More recently, Feuz and Cook [11] investigated heuristics methods to find an optimal many-to-one mapping through the use of greedy search and genetic algorithms. For these heuristics methods, reported results show greater accuracy and average recall than a manual feature mapping method; nevertheless, the gigantic search space makes the algorithms computationally expensive.

Among those approaches not limited to sensor networks, Zhou et al. [12] formalized an optimization algorithm to learn feature remapping with pivots. While other features are domain-dependents, pivots are described as features that can be commonly shared across domains; therefore, pivots can be used as the guides to transfer domain-specific features. Zhou et al. [13] introduced an algorithm to learn a sparse feature transformation for heterogeneous domain adaptation which allow them to transfer knowledge among support vector machines. Sukhija et al. [14] extended this work using random forests to estimate label distributions that are used as pivots across domains. They compared the models with a baseline of classification task on CASAS datasets [6] and showed a significant decrease of the mean error. However, sparse feature transformation is a supervised method, hence it requires labels from target domain which is often not suitable for real applications.

While *feature remapping* approaches heterogeneous domain adaptation with a direct mapping between domains, *latent space transformation* focuses more on constructing a projection space shared across domains. The common principle is to learn to (1) project domain-specific data to a specific latent space, (2) use the labels from the source domain to perform a task on the latent space, (3) project data from target domains in the latent space, and finally (4) perform the desired task, e.g. classification, on latent projections. The main challenge resides in the construction of the latent space projection. Shi et al. [15] introduced heterogeneous spatial mapping, an unsupervised method, to align source and target domains using spectral properties, sample selection and Bayesian modelling of output spaces. Another example is Wang et al. [16], which provides an algorithm to construct the projection that preserves the local sample neighborhood in source manifolds while letting similar samples from different manifolds be the neighborhoods in the projection space.

Recent advances in deep learning bring new robust frameworks to approach *latent space transformation*. Zhuang et al. [17] used deep autoencoders to create latent embeddings of domain-specific features. The Kullback-Leibler divergence between embedding distributions is minimized, which allows embedding spaces to converge to a shared space. Wang et al. [18] also leveraged autoencoders to construct a shared feature space, using Maximum Mean Discrepancy and a manifold alignment term to preserve the local geometric structure of data while reducing differences in latent features distributions. Recently, Haeusser et al. [9] introduced *associative domain adaptation*, a method taking advantage of labels in the source domain to create clustered representations of data from source and target domains. This method requires to have labels from the source domain but has the benefit of preserving local structure and thus can create latent spaces with consistent clusters.

In our work, we propose a *latent space transformation* framework based on a new method that performs representation learning on the sensor events modelled as graphs. Representation learning for graphs has been investigated in several works. In particular, Kipf et al. [19] introduced a two kinds of Graph Autoencoder (GAE) using an inner product between latent variables to reconstruct the adjacency matrix of graphs. Wang et al. [20] leveraged linear graph convolutional networks and GAE to propose a new autoencoder called Marginalized Graph Autoencoder (MGAE). This model corrupts input nodes representation by randomly turning some components to zero. For models using graph convolutional networks to acquire a representation of node features and adjacency, the major issue resides in the variable size of the latent representation, which depends on the number of nodes in the graph. Here, we present a novel GAE architecture that leverages *associative domain adaptation* and *differentiable pooling* to acquire a structure independent, domain invariant representation of graphs.

## 3   Structure-Independent Graph Autoencoder for HSNs

The structure-independent model introduced here enables usage of machine learning across HSNs. It relies on graph convolutional networks, differentiable pooling and associative domain adaptation loss to learn to project graph representations into a shared latent space. We first introduce the notation and prerequisite concepts in Section 3.1, before presenting our framework in Section 3.2.

### 3.1   Prerequisites

**Notations.** Let $\mathcal{G} = \{\mathcal{V}, \mathcal{E}, \mathcal{W}\}$ be a graph with a set of $N$ nodes $v_i \in \mathcal{V}$, a set of edges $e = (v_i, v_j) \in \mathcal{E}$, and edge weights $\mathcal{W}(e) \in \mathbb{R}, \forall e \in \mathcal{E}$. Given a node ordering for $\mathcal{G}$,

5we use the notation $\mathcal{G} = \{X, A\}$, where $A \in \mathbb{R}^{N \times N}$ represents the *adjacency matrix* and each node $v_i$ is represented by a vector of $F$ attributes and collected in a *matrix of node embeddings*, $X \in \mathbb{R}^{N \times F}$. In the following, the notation $\tilde{A}$ is used to refer the *normalized adjacency matrix* $\tilde{A} = \bar{D}^{-1/2} \bar{A} \bar{D}^{-1/2}$, where $\bar{A} = A + I_N$, $I_N$ is the $N \times N$ identity matrix and $D$ is the degree matrix, $\bar{D}_{ij} = \sum_j \bar{A}_{ij}$ if $i = j$, 0 otherwise.

**Graph Convolutional Networks.** Graph Convolutional Networks (GCNs) [7] learn a *convolution kernel* that uses features of neighbors nodes to perform a potentially non-linear mapping of nodes embeddings from their initial representation to a different space. A GCN can be compounded of several layers stacked together. Let $d_l$ be the number of features at layer $l$. In a $L$-layer(s) GCN, a single layer transforms nodes representation from $\mathbb{R}^{N \times d_l}$ to $\mathbb{R}^{N \times d_{l+1}}$ by applying:

$$X^{(l+1)} = \sigma(\tilde{A} X^{(l)} W^{(l)}) \tag{1}$$

where $W^{(l)} \in \mathbb{R}^{d_l \times d_{l+1}}$ is a matrix of trainable parameters, $\sigma(\cdot)$ is an activation function, $X^{(l)} \in \mathbb{R}^{N \times d_l}$ represents the *matrix of nodes embedding* at layer $l$.

A common application is to use the last layer nodes representation as an input for a differentiable classifier, e.g. a Multi-Layer Perceptron (MLP). The entire model can be efficiently end-to-end trained to classify graph-structured data. In the following, we use the notation $\text{GCN}(A, X) = \tilde{A} X W$ to denote the application of graph convolution on a graph $\mathcal{G} = \{A, X\}$ with a linear activation function.

**Differentiable Pooling.** As presented in Eq. 1, a layer of a GCN transforms the nodes representation without modifying the graph structure. Ying et al. [8] introduced a differentiable pooling method to modify the graph structure and to learn hierarchical representations of graphs. Pooling layers, also called DIFFPOOL layers, are interleaved between some layers of a graph neural network to coarse the graph representation. A DIFFPOOL layer transforms the original graph structure by clustering nodes. At the $l$-th layer, we denote the input matrix of embeddings as $X^{(l)} \in \mathbb{R}^{n_l \times d_l}$ and the input adjacency matrix as $A^{(l)} \in \mathbb{R}^{n_l \times n_l}$. The DIFFPOOL layer constructs a new adjacency matrix, $A^{(l+1)} \in \mathbb{R}^{n_{l+1} \times n_{l+1}}$, and a new matrix of embeddings $X^{(l+1)} \in \mathbb{R}^{n_{l+1} \times d_{l+1}}$, such as $(A^{(l+1)}, X^{(l+1)}) = \text{DIFFPOOL}(A^{(l)}, X^{(l)})$.

To perform the transformation between graph structures, a DIFFPOOL layer learns to construct an *assignment matrix* which defines how the nodes in the input graph are *assigned* to the nodes in the output graph. Given $S^{(l)} \in \mathbb{R}^{n_l \times n_{l+1}}$ an assignment matrix for the $l$-th layer, the transformation applied to the graph structure is:

$$\begin{aligned} X^{(l+1)} &= S^{(l)^T} Z^{(l)} \\ A^{(l+1)} &= S^{(l)^T} A^{(l)} S^{(l)} \end{aligned}$$

Two GCNs can be used to construct a new matrix of node embeddings $Z^{(l)}$ and the assignment matrix $S^{(l)}$:

$$Z^{(l)} = \sigma\left(GCN_{l,Z}(A^{(l)}, X^{(l)})\right)$$
$$S^{(l)} = softmax(\text{GCN}_{l,P}(A^{(l)}, X^{(l)}))$$

where $\sigma(\cdot)$ is an activation function and $softmax(\cdot)$ is the row-wise function defined as $softmax_i(x) = \frac{\exp(x_i)}{\sum_j \exp(x_{i,j})}$.

We emphasize that both the matrix of node embeddings and the adjacency matrix are constructed using GCNs with fixed output dimensions. Hence, DIFFPOOL layers learn to project into a *latent space with fixed dimensions* the graph representations with potentially different adjacencies or number of nodes.

**Associative Domain Adaptation.** Introduced by Haeusser et al. [9], associative domain adaptation aims at acquiring consistent projection domains from statistically different source domains. Given $Z_i^s$ and $Z_j^t$ as the embeddings respectively from source and target domains, associative domain adaptation [9] computes a similarity measure matrix using the dot product $M_{i,j} = \langle Z_i^s, Z_j^t \rangle$. Considering the set of source and target representations as a bipartite graph, this similarity measure matrix is used to estimate the transition probability from node embedding $Z_i^S$ to node embedding $Z_j^T$:

$$P_{ij}^{st} = \mathbb{P}(Z_j^t | Z_i^s) = \frac{\exp(M_{ij})}{\sum_{j'} \exp(M_{ij'})}$$

From the transition probability, the authors derive two losses which will be minimized during the training step. The first one, *walker loss*, forces round trips within the same class to have the same probability by expressing $\mathcal{L}_{walker} = H(T, P^{sts})$, where $H$ is the cross-entropy function, $P_{ij}^{sts} = (P^{st}P^{ts})_{ij}$ is the probability of a first order round trip and $T_{ij} = 1/|Z_i^s|$ if $i$ and $j$ have the same class, 0 otherwise. The second loss is a *visit loss* that forces a uniform probability of visiting target examples, $\mathcal{L}_{visit} = H(V, P^{visit})$, where $V_j = |Z^t|$ and $P_j^{visit} = \sum_i P_{ij}^{st}$.

The overall loss $\mathcal{L}_{assoc}$ is the weighted sum of walker and visit losses:

$$\mathcal{L}_{assoc} = \beta_{\text{walker}} \mathcal{L}_{\text{walker}} + \beta_{\text{visit}} \mathcal{L}_{\text{visit}} \qquad (2)$$

A better domain adaptation is achieved if the second weight $\beta_{visit}$ is decreased when label distribution strongly varies between source and target domains.

### 3.2 Framework Presentation

The number of sensors, their types and their layouts vary among sensor networks. Intuitively, we want to exploit DIFFPOOL layers ability to build latent spaces of fixed dimensions together with the capacity of associative domain adaptation to acquire domain invariant representations. The key idea is to construct structure-independent representations of HSNs states usable as inputs for structure-dependent machine learning algorithms. First, we construct a simple graph representation that captures some structural and semantic knowledge about sensors types and layout in a smart home. Next, we present our structure-independent autoencoder. Finally, we introduce the overall objective function enabling domain-invariant representation learning.

**Simple Graph Representation for Sensor Networks in Smart Homes.** We focus here on simple sensors attached to appliances in smart homes and producing 1-dimensional measures. The minimal semantic information necessary to work with these sensors is their locations in the smart home, e.g. kitchen or bathroom, and the quantity/state they measure, e.g. room temperature or door state. Adjacency between locations is also required. Given semantic information and an ordering for sensors, we define the following rule to build adjacency matrices for smart homes:

$$A_{ij} = \begin{cases} 1 & \text{if sensors } i \text{ and } j \text{ are in the same location} \\ \frac{1}{2} & \text{if } i \text{ and } j \text{ are in adjacent locations} \\ 0 & \text{otherwise} \end{cases}$$

As this design of adjacency matrices is arbitrary, we discuss its pertinence and influence in Section 5.3. We also need to build a simple node representation of sensors states at a particular time. We first define an ordered set of generic sensor types representing the quantities or states measured by sensors. As we are informed of sensor state changes through events, we need to gather sensor events to construct features that are representatives of the overall sensor network states at a particular time. We use fixed-window sampling to gather sensors events in non-overlapping windows with duration $T_{window}$. For a window $[t, t + T_{window}[$, we construct the sensor network representation $X_t$ by counting the number of times each sensor fired, i.e. if the $i$-th sensor fires $n$ times in the window and is of the $j$-th type, then $(X_t)_{i,j} = n$. This guarantees that sensors from different datasets have the same features: each line of the matrix $X_t$ is a one-hot vector representing the sensor's type and number of firings within the window.

**Encoding the Graph Representation.** Like other autoencoders, our model consists of an encoder part, that learns to project the data in the latent space, and a decoder part, that reconstructs input data from the latent representation. The encoder is compounded of at least one differentiable pooling layer which uses node features to perform the projection in a fixed-size latent space. The latent representation is used as input for the model learning to perform the application task, e.g. classifying states.

For a set of graph representations with the same feature representation but possibly different number of nodes, we construct the latent representations by applying our encoder model, $Z = \text{ENCODER}(A, X)$. For all positive non-zero number of nodes in the graph representation, the matrix of encoded embeddings satisfies $Z \in \mathbb{R}^{N_H \times D_H}$, where $N_H$ and $D_H$ are hyperparameters of the ENCODER model. Hence, we construct different graph representations for different sensor networks, and encode them in a latent space of custom dimensions. In our experiments, we use a simple encoder model defined as follows:

$$H = \text{ReLU}\big(\text{GCN}_H(A, X)\big)$$
$$A_{enc}, Z = \text{DIFFPOOL}(A, H)$$

where the first layer uses the Rectifier Linear Unit, $\text{ReLU}(x) = \max(0, x)$, as activation function. The DIFFPOOL layer uses GCN, as presented in Eq. 1 with the *hyperbolic tangent* as activation function (denoted $\sigma(\cdot)$ in Eq. 1).

**Training the Graph Autoencoder.** From a matrix of encoded embeddings $Z$, the decoder reconstructs an approximate node representation $\hat{X} = \text{DECODER}(Z)$. In our experiments, we use the following two-layers decoder model:

$$H_{\text{dec}} = \text{ReLU}\left(\text{GCN}_{\text{dec},1}(A, Z)\right)$$
$$\hat{X} = \text{ReLU}\left(\text{GCN}_{\text{dec},2}(A, H_{\text{dec}})\right)$$

The overall loss used for training is compounded of several losses with complementary objectives. To train the model as an autoencoder, we define a reconstruction loss $\mathcal{L}_{rec}$ that measures the distance between the input matrix of node embeddings $X$ and the reconstructed representation $\hat{X}$. In our experiments we used the Euclidian distance between $X$ and $\hat{X}$, generally known as L2 loss, although many different losses can be used.

A second loss is associated to the assignment matrix of the pooling layer. Ying et al. [8] introduced this loss to ensure that the pooling layer finds a nearly optimal clustering with well-defined nodes assignments. It is expressed as:

$$\mathcal{L}_{pool} = \|A - SS^T\|_F + \frac{1}{n}\sum_{i=1}^{n} H(S_i)$$

where $H$ is the entropy function and $S_i$ the $i$-th row of the assignment matrix $S$.

The last loss is the associative domain adaptation loss from Eq. 2, $\mathcal{L}_{assoc}$, ensuring the consistency of the shared latent space across domains. The overall loss for the graph autoencoder training step is $\mathcal{L}_{ae} = \mathcal{L}_{rec} + \alpha_{\text{pool}}\mathcal{L}_{pool} + \alpha_{\text{assoc}}\mathcal{L}_{assoc}$. The set of model parameters is iteratively optimized to decrease this global loss.

## 4 Experimental Setup

We use the previously introduced encoder model to project HSNs states in a shared latent space. The machine learning model to transfer across HSNs is then designed and trained on the latent projections made by the encoder model. The overall model, composed of both the encoder and the model to transfer, is trained and used across HSNs. We focus here on using our encoder model to transfer activity recognition models across HSNs deployed in smart homes. Section 4.1 introduces the datasets used in our experimental setup and the preprocessing applied to data. In Section 4.2, we present the models and hyperparameters used in our experiments.

### 4.1 Datasets and Preprocessing

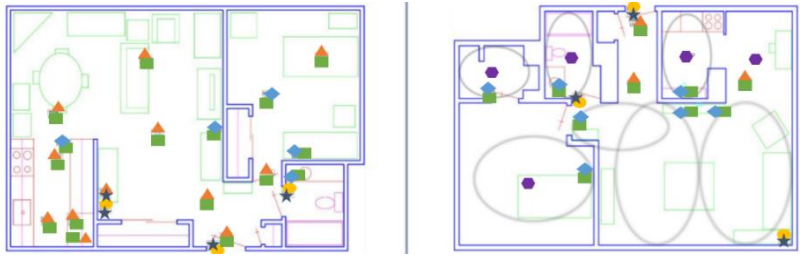

**Fig. 2.** Example of heterogeneous sensor layouts in the smart homes used to collect *hh101* (left) and *hh103* (right). Different shapes represent different sensors types.

**Datasets.** We use *hh101* to *hh105*, a set of Human Horizon (HH) datasets collected by CASAS [6] on several smart homes with single occupants. The heterogeneity of layouts among smart homes, depicted in Fig. 2, makes these datasets good data sources to evaluate domain adaptation methods. Each dataset provides raw sensor data with annotated activities. The sensors deployed in the smart homes are either binary sensors attached to appliances or real-valued sensors such as temperature or brightness sensors. Omitting the battery sensors, we count a total of 6 different sensor types. In the dataset, accurate activity annotations result in many labels, which leads to unbalanced class representations. We gather similar classes together to form 13 labels used for activity recognition. In Table 1, we provide a description of the datasets with the number of sensors, events, activities as well as the sensor types and activity labels used for evaluation.

**Table 1.** Description of Human Horizon datasets from CASAS [6]

| Dataset | hh101 | hh102 | hh103 | hh104 | hh105 |
|---|---|---|---|---|---|
| Nb. of sensors | 40 | 64 | 37 | 70 | 53 |
| Nb. of events | 321,645 | 407,583 | 164,908 | 478,003 | 222,591 |
| Sensor types | Door switch, Light switch, Light, Wide area motion, Temperature, Motion ||||| 
| Activity clusters | Unclassified, Personal Hygiene, Cooking, Eating, Working, Entering/Leaving Home, House Keeping, Taking Medicine, Washing Dishes, Toilet, Relaxing, Exercising, Other |||||

**Preprocessing.** Using smart homes layouts as depicted in Fig. 2, we apply the method presented in Section 3.2. to create simple graphs representations of smart homes. We use a window length of $T_{window} = 180$ seconds to create graph representations. The activity label associated with a window is set to the activity performed during the longest part of the time window. The six different sensor types used to build representations of sensors states are presented in Table 1.

### 4.2 Experimental Setup

We use simple and well-known machine learning algorithms to perform activity recognition: decision trees (DT), *k*-nearest neighbors (KNN) and multilayer perceptrons (MLP). Our DT classifiers use Gini impurity to determine the quality of splits. We always choose the best split to recursively divide a set of training samples in two subsets, with a maximum number of leaves set to 500. Our KNN models classify a point by returning the value of the $k = 3$ nearest neighbors. We construct two-layers MLP with 64 hidden units and *hyperbolic tangent* activation in the first layer, and 13 output units with *softmax* activation in the second layer. We refer to these baseline models as **DT**, **KNN** and **MLP**. We use a set of preprocessed data with the corresponding activity labels to train three activity classifiers without encoder. Data are dispatched between a training and a testing set. The MLP model is trained to minimize the cross-entropy loss on sample labels, using a batch size of 64 and Adam optimizer [21] with a learning rate of $5 \times 10^{-4}$. We add an L2 regularizer loss with a weight of $1 \times 10^{-4}$ to avoid overfitting.

We created our GAE model with the architecture described in Section 3. The first GCN layer transforms nodes representations from the $F = 6$ shared features, i.e. the number of sensor types, to 32 latent features. Next, the DIFFPOOL layer performs a projection in a latent space of fixed dimensions $N_H \times F_H$, with $N_H = 64$ and $F_H = 16$. We empirically selected appropriate loss weights, resulting in a pooling loss weight $\alpha_{pool} = 1 \times 10^{-4}$ and an association loss weight $\alpha_{assoc} = 0.1$, with $\beta_{walker} = 1$ and $\beta_{visit} = 0.3$. The model is trained with a batch size of 64 using Adam optimizer and a learning rate of $5 \times 10^{-4}$. We trained the model on 100 epochs, using early stopping with a patience of 10, i.e. the training stops if the validation loss does not decrease for 10 consecutive epochs. We used two different datasets to train the autoencoder. Training samples are then encoded with our model and used to train new classifiers with the same hyperparameters as the baselines. We refer to the combination of the GAE with DT (resp. KNN and MLP) models as **GAE+DT** (resp. **GAE+KNN**, **GAE+MLP**). We take advantage of MLP differentiability by end-to-end fine-tuning the GAE+MLP model with a learning rate of $5 \times 10^{-5}$.

In the following, we use two metrics to evaluate our models. The first one is the well-known *F1-score*, calculated for a classification task as the harmonic average of precision and recall. We report the average and 95% confidence interval of the F1-score collected on 10 experiments. Due to strong variations between baseline scores, an absolute metric like F1-score is heterogeneous across models. We introduce a second metric, called *relative score*, to compare on a common scale models' transferability across datasets. Given a model composed of a classifier trained with a graph autoencoder, the relative score for a target dataset is the ratio between the average F1-score obtained by the model and the average F1-score achieved by the classifier baseline, expressed in percent. This score quantifies how the model performs *relatively* to its baseline.

11## 5 Results and Discussion

Our framework aims at applying the same machine learning model to classify activities using sensor events from several HSNs deployed in smart homes. The activity classifier model can be trained using available labels and data from one or several datasets. We focus on evaluating the performances of our model in a simple case where the activity classifier model is trained on a single *source* dataset, representing data from a *source* smart home layout. The model is used to classify activities on a *target* dataset, representing a *target* smart home layout. Results for transfer of activity classifiers models for different source → target datasets pairs are reported and discussed in Section 5.1. Due to the use of associative domain adaptation loss, the encoder model requires *data* from the target dataset in addition to *data and labels* from the source dataset. In Section 5.2, we quantify the relative score for activity recognition depending on the quantity of available data for target dataset to evaluate how fast activity classifier will adapt to unseen HSNs. In Section 5.3, we discuss the pertinence of the arbitrary adjacency matrix design introduced in Sections 3.2 and evaluate the resilience of our framework to suboptimal adjacency matrix design by comparing transfer results for different kinds of adjacency matrix.

### 5.1 Results for Transfer of Machine Learning Models across HSNs

We evaluate the ability of the encoder to transfer the knowledge acquired on a source HSN to a target HSN. GAE+DT, GAE+KNN and GAE+MLP models are trained using *data and labels* from a source dataset but *only data* from a target dataset. This experiment represents the practical case where labeled data are available only for the source sensor network, but some unlabeled data have been collected on the target sensor network. The results collected for some source → target pairs are reported in Table 2.

**Table 2.** F1-score and relative score of GAE models evaluated on different datasets pairs.

| Model | | hh101 → hh102 | hh102 → hh101 | hh103 → hh105 | hh104 → hh103 | hh105 → hh104 |
|---|---|---|---|---|---|---|
| **GAE +DT** | *F1-score* | 56.6 (0.9) | 57.8 (1.7) | 62.6 (1.0) | 52.3 (1.9) | 45.3 (0.6) |
| | *Relative score* | **83.7** | 75.7 | 80.7 | 75.8 | 69.4 |
| **GAE +KNN** | *F1-score* | 55.6 (0.3) | 54.5 (0.6) | 62.2 (0.6) | 50.5 (0.8) | 44.1 (0.3) |
| | *Relative score* | **84.8** | 74.1 | 81.6 | 74.6 | 70.3 |
| **GAE +MLP** | *F1-score* | 57.8 (0.7) | 57.1 (2.5) | 65.1 (0.7) | 48.1 (0.9) | 45.8 (0.5) |
| | *Relative score* | **104.7** | 85.1 | 92.7 | 81.3 | 81.2 |

Results show that:

- GAE+DT reaches a mean F1-score of 54.9% on the presented source-target pairs. The model performs differently among datasets pairs, with relative score reaching up to 83.7% of the DT baseline score for the pair hh101 → hh102 and down to 69.4% of the DT baseline score for the hh105 → hh104. In average, the GAE+DT model trained on a source dataset achieves 77.0% of the DT baseline score on the target dataset.
- The GAE+KNN model achieves an average relative score of 77.1% of the KNN baseline score, with an average F1-score of 53.4%. Like the GAE+DT model, the higher relative score, 84.8%, is reached for the pair hh101 → hh102. The GAE+KNN model obtains its lower relative scores for hh102 → hh101 and hh105 → hh104, with respectively 74.1% and 70.3% of the KNN baseline score.
- The GAE+MLP model reaches an average relative score of 88.7% of the MLP baseline score, outperforming GAE+DT and GAE+KNN. Once again, the highest relative score is achieved for the pair hh101 → hh102, with a F1-score of 57.8% representing 104.7% of the MLP baseline score. Like GAE+DT and GAE+KNN, the model achieves its lower relative scores for hh104 → hh103 and hh105 → hh104 with respectively 81.3% and 81.2% of the baseline score.

All three models perform differently across source-target pairs. However, we observe that the scores are consistent among models: all of the models achieved their highest relative score on hh101 → hh102 and their second highest relative score on hh103 → hh105. GAE+DT and GAE+KNN reached their two lowest relative score on hh105 → hh104 and hh102 → hh101, while GAE+MLP achieved its lowest relative scores on hh105 → hh104 and hh104 → hh103. These results give the impression that the difficulty of the transfer task varies among source-target datasets pairs, causing different relative scores. There is no explanation on the observed results difference and further investigations would be required to identify the causes underlying this difference.

In our experiments, the GAE+MLP model always outperformed the GAE+DT and GAE+KNN models in terms of relative score. This fact is due to two reasons. First, the MLP baseline is a differentiable model with more hyperparameters than the DT or KNN baselines. Hence, despite our efforts to select appropriate hyperparameters, the MLP baseline is more likely than DT or KNN to achieve suboptimal results. Consequently, the selected hyperparameters can be more appropriate for GAE+MLP than for the MLP baseline, resulting in a greater relative score for the GAE+MLP model. The second reason behind GAE+MLP high scores is fine-tuning. While the encoder and the classifier are only trained separately for GAE+DT and GAE+KNN, GAE+MLP benefits from an end-to-end fine-tuning of the entire model. Hence, the GAE is optimized to help the MLP achieving better activity recognition results. This fine-tuning step helps GAE+MLP to achieve higher relative scores than GAE+DT or GAE+KNN.

## 5.2 Evaluation of Deployment Speed.

Some data from the target dataset is required to apply associative domain adaptation in our framework. We evaluate how the number of data collected on the target environment influences the average relative score of our models. Relative scores for different number of data from target datasets are reported on Fig. 3. As we use a fixed-size window of three minutes, each data point represents three minutes, e.g. 20 data points represent one hour.

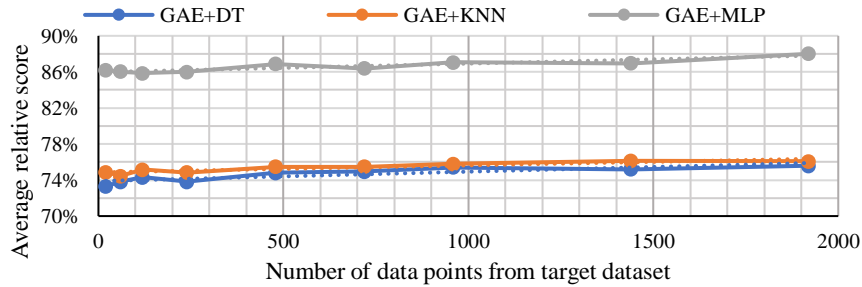

**Fig. 3.** Evaluation of the Adaptation Speed: Average Relative Score for Different Number of Data Points from Target Dataset

We observe that the average relative score will be slightly increased if there is more data from the target HSN. This is the main purpose of using associative domain adaptation. More surprisingly, the model still performs quite decently with a low number of data from target environment. For instance, the GAE+DT model achieves 73.2% average relative score with only 20 data points from the target dataset, which represents only 3.8% less than the relative score obtained in Section 5.1 with every available data points. Hence, if more data points from the target HSNs helps transferring the machine learning model across HSNs, the model still performs decently when only a few data from target HSN is available.

### 5.3 Influence of the Adjacency Matrix Design

In Section 3.2, we arbitrarily design adjacency matrices to express intuitive adjacency between sensors given the layout of an HSN in a smart home. However, this arbitrary design might be suboptimal. Adjacency matrix is a fundamental parameter in graph convolutional layers and strongly influences the way DIFFPOOL layers cluster nodes. Though it goes beyond our scope to investigate which design of adjacency matrix would best represent sensors layouts in smart homes, we still want to evaluate the influence of the adjacency matrix design.

The previously evaluated model uses the adjacency matrix design from Section 3.2, which we call *Default* in the following. The transfer learning results obtained with three additional, different adjacency matrices are compared against the results obtained with our previously evaluated model. The first kind of adjacency matrix (*Identity*) is a zero matrix, which results in the identity matrix after symmetric normalization. The second kind of adjacency matrix (*FC-U*) represents an unweighted fully connected graph. The third kind of adjacency matrix (*FC-W*) is based on the adjacency matrix design introduced in Section 3.2 but turned into a fully connected graph by setting the weight between sensor in different non-adjacent locations to 0.1 instead of 0.

For these different adjacency matrix designs, we trained and evaluated GAE+DT models on pairs of source-target datasets. We choose to report results for the GAE+DT model as it achieves decent performances within a limited time and requires only a few hyperparameters tuning. Relative scores obtained for the four different adjacencies designs are presented in Table 3.

We see that the proposed adjacency matrix design outperforms other kinds of adjacency matrices on every source-target dataset pair, except for hh101→hh102. Average relative scores are 77.0% for the *Default* design, 75.0% for the *Identity* design, 74.6% for the *FC-U* design and 75.9% for the *FC-W* design. Hence, if the choice of adjacency in the graph structure helps achieving better transfer results, its influence seems limited. In our case, the worst choice was to represent the sensor network as an unweighted fully connected graph. However, this poor choice of a graph structure only resulted in a loss of 2.4% of the average relative score. We conclude that the influence of the adjacency matrix is limited, and thus our framework is relatively resilient to suboptimal designs of the graph structures representing the HSNs.

**Table 3.** Relative Scores for Different Kinds of Adjacency Matrices

| Kind of Adjacency | hh101 → hh102 | hh102 → hh101 | hh103 → hh105 | hh104 → hh103 | hh105 → hh104 |
|---|---|---|---|---|---|
| Default | 83.7 | **75.7** | **80.7** | 75.8 | 69.4 |
| Identity | **86.5** | 74.1 | 76.7 | 68.0 | 69.4 |
| FC-U | 84.6 | 72.1 | 78.9 | 68.3 | 69.0 |
| FC-W | 83.6 | 74.3 | 78.7 | 75.5 | 67.4 |

## 6 Conclusion and future work

When designing an application based on machine learning and targeting sensor networks, the heterogeneity of sensor layouts and types is a common issue. Indeed, the data collection and model training effort must be repeated for each new sensor network. A practical example is activity recognition is smart homes: a new machine learning model must be created and trained for each new smart home. In this paper, we propose a new method leveraging graph representation learning with autoencoders to build latent representations independent from the type or layout of sensors across HSNs.

We introduce a simple graph representation of the state of HSNs in smart home to enable the use of graph autoencoders. Our model relies on differentiable pooling and GCNs to project the representations of events in HSNs to latent spaces of fixed dimension. We then train our model as an autoencoder with an associative domain adaptation term encouraging to share the latent spaces between HSNs. Activity recognition models are trained to classify activities from the shared latent space, which makes the models independent to the structure of target HSNs. We focus here on transferring activity recognition models across HSNs deployed in smart homes. We use CASAS datasets to train baseline activity classifiers models based on decision trees, k-nearest neighbors and multilayer perceptron. After building the graph representation for events in HSNs, we train our graph autoencoder model to create the shared latent representations of events in HSNs. Activity classifiers models are trained on top of the shared latent spaces and compared with their respective baselines.

The obtained results imply that our framework allows to train an activity recognition model based on DT (resp. KNN, MLP) on data from a source smart home with known activity labels and to apply the model to data from a target smart home without activity labels, with an F1-score representing in average 77.0% (resp. 75,0%, 88,7%) of the baseline score on the target smart home. Moreover, the models require only a few data from target HSN to achieve decent performances. In addition, we assume that the proposed adjacency matrix design is suboptimal and we evaluate the influence of the graph structure on the performances of our model. It appears that the structure of the graphs representing the state of HSNs have a quite limited influence on the results. Hence, our model still performs decently with suboptimal graph representations of HSNs.

Enabling the usage of graph neural networks for HSNs opens the range of applicable solutions. The key idea is to instill prior structural knowledge into a machine learning algorithm, with the intuition that models can exploit this structural knowledge to acquire structure-independent representations of sensor events. However, as we presented in Section 3.2, the design of an adjacency matrix, and hence the choice of a structural representation, is a complex but fundamental step. Recent work on graph neural networks by Schlichtkrull et al. [22] focused in instilling semantic knowledge into multi-graphs and proposed frameworks to perform semantic-relation-wise graph convolution. Hence, each layer can acquire a convolution kernel for type of semantic relationship. In these models, adjacency matrices are relation-wise, meaning that we can express different kinds of semantic relationships between sensors. Hence, in the context of cognitive IoT, we can exploit the ubiquity of semantic annotations to create complex structural representations of sensor networks. For instance, sensor adjacency in smart environment could represent that sensors are physically adjacent, but also that they share the same sensor type or are deployed in the same type of room. To summarize, semantic GCNs could take advantage of rich semantic annotations to enhance machine learning applications.

## **Acknowle**d**gment**

**References**

1. Suryadevara, N.K., Mukhopadhyay, S.C., Wang, R., Rayudu, R.K.: Forecasting the behavior of an elderly using wireless sensors data in a smart home. Eng. Appl. Artif. Intell. 26, 2641–2652 (2013). doi:10.1016/J.ENGAPPAI.2013.08.004
2. Orpwood, R., Adlam, T., Evans, N., Chadd, J., Self, D.: Evaluation of an assisted-living smart home for someone with dementia. J. Assist. Technol. 2, 13–21 (2008)
3. Lotfi, A., Langensiepen, C., Mahmoud, S.M., Akhlaghinia, M.J.: Smart homes for the elderly dementia sufferers: Identification and prediction of abnormal behaviour. J. Ambient Intell. Humaniz. Comput. 3, 205–218 (2012). doi:10.1007/s12652-010-0043-x
4. Barth, J., Klucken, J., Kugler, P., Kammerer, T., Steidl, R., Winkler, J., Hornegger, J., Eskofier, B.: Biometric and mobile gait analysis for early diagnosis and therapy monitoring in Parkinson's disease. In: 2011 Annual International Conference of the IEEE Engineering in Medicine and Biology Society. pp. 868–871. IEEE (2011)
5. Chiang, Y., Lu, C.-H., Hsu, J.Y.-J.: A Feature-Based Knowledge Transfer Framework for Cross-Environment Activity Recognition Toward Smart Home Applications. IEEE Trans. Human-Machine Syst. 47, 310–322 (2017). doi:10.1109/THMS.2016.2641679
6. Cook, D.J., Crandall, A.S., Thomas, B.L., Krishnan, N.C.: CASAS: A Smart Home in a Box. Computer (Long. Beach. Calif). 46, 62–69 (2013). doi:10.1109/MC.2012.328
7. Kipf, T.N., Welling, M.: Semi-Supervised Classification with Graph Convolutional Networks. arXiv Prepr. arXiv1609.02907. (2016)
8. Ying, R., You, J., Morris, C., Ren, X., Hamilton, W.L., Leskovec, J.: Hierarchical Graph Representation Learning with Differentiable Pooling. In: Advances in Neural Information Processing Systems. pp. 4800–4810 (2018)
9. Haeusser, P., Frerix, T., Mordvintsev, A., Cremers, D.: Associative Domain Adaptation. Proc. IEEE Int. Conf. Comput. Vis. 2017-Octob, 2784–2792 (2017). doi:10.1109/ICCV.2017.301
10. Hu, D.H., Yang, Q.: Transfer learning for activity recognition via sensor mapping. In: IJCAI International Joint Conference on Artificial Intelligence. pp. 1962–1967 (2011)
11. Dillon Feuz, K., J. Cook, D., Feuz, K.D., Cook, D.J.: Heterogeneous transfer learning for activity recognition using heuristic search techniques. (2014)
12. Zhou, G., He, T., Wu, W., Hu, X.T.: Linking Heterogeneous Input Features with Pivots for Domain Adaptation.
13. Zhou, J.T., Tsang, I.W., Pan, S.J., Tan, M.: Heterogeneous Domain Adaptation for Multiple Classes. (2014)
14. Sukhija, S., Krishnan, N.C., Singh, G.: Supervised Heterogeneous Domain Adaptation via Random Forests. In: International Joint Conferences on Artificial Intelligence. pp. 2039–2045 (2016)
15. Shi, X., Liu, Q., Fan, W., Yu, P.S., Zhu, R.: Transfer learning on heterogenous feature spaces via spectral transformation. Proc. - IEEE Int. Conf. Data Mining, ICDM. 1049–1054 (2010). doi:10.1109/ICDM.2010.65
16. Wang, C., Mahadevan, S.: Manifold alignment without correspondence. (2009)
17. Zhuang, F., Cheng, X., Luo, P., Pan, S.J., He, Q.: Supervised representation learning: Transfer learning with deep autoencoders. In: IJCAI International Joint Conference on Artificial Intelligence. pp. 4119–4125 (2015)
18. Wang, X., Ma, Y., Cheng, Y., Zou, L., Rodrigues, J.J.P.C.: Heterogeneous domain adaptation network based on autoencoder. J. Parallel Distrib. Comput. 117, 281–291 (2018). doi:10.1016/j.jpdc.2017.06.003
19. Kipf, T.N., Welling, M.: Variational Graph Auto-Encoders. (2016)

20. Wang, C., Pan, S., Long, G., Zhu, X., Jiang, J.: MGAE: Marginalized Graph Autoencoder for Graph Clustering. doi:10.1145/3132847.3132967
21. Kingma, D.P., Ba, J.: Adam: A Method for Stochastic Optimization. (2014)
22. Schlichtkrull, M., Kipf, T.N., Amsterdam pbloem, V., Rianne van den Berg,  vunl, Titov, I., Welling, M.: Modeling Relational Data with Graph Convolutional Networks Peter Bloem. In: European Semantic Web Conference. pp. 593–607 (2017)